\documentclass[conference]{IEEEtran}
\IEEEoverridecommandlockouts
\usepackage{cite}
\usepackage{amsmath,amssymb,amsfonts}
\usepackage{hyperref}
\usepackage{algorithmic}
\usepackage{graphicx}
\usepackage{textcomp}
\usepackage{makecell}
\usepackage{xcolor}
\usepackage{cprotect}
\usepackage{multirow}
\def\BibTeX{{\rm B\kern-.05em{\sc i\kern-.025em b}\kern-.08em
    T\kern-.1667em\lower.7ex\hbox{E}\kern-.125emX}}
\begin{document}

\title{Play Style Identification Using Low-Level Representations of Play Traces in \textit{MicroRTS}\\

\thanks{Short Paper. This work was supported by the EPSRC Centre for Doctoral Training in Intelligent Games \&
Game Intelligence (IGGI) EP/S022325/1}
}

\author{\IEEEauthorblockN{Ruizhe Yu Xia}
\IEEEauthorblockA{\textit{Queen Mary University of London}\\
London, United Kingdom \\
r.yuxia@qmul.ac.uk}

\and
\IEEEauthorblockN{Jeremy Gow}
\IEEEauthorblockA{\textit{Queen Mary University of London}\\
London, United Kingdom \\
jeremy.gow@qmul.ac.uk}
\and
\IEEEauthorblockN{Simon Lucas}
\IEEEauthorblockA{\textit{Queen Mary University of London}\\
London, United Kingdom \\
simon.lucas@qmul.ac.uk}
}

\maketitle

\begin{abstract}
Play style identification can provide valuable game design insights and enable adaptive experiences, with the potential to improve game playing agents. Previous work relies on domain knowledge to construct play trace representations using handcrafted features. More recent approaches incorporate the sequential structure of play traces but still require some level of domain abstraction. In this study, we explore the use of unsupervised CNN-LSTM autoencoder models to obtain latent representations directly from low-level play trace data in \textit{MicroRTS}. We demonstrate that this approach yields a meaningful separation of different game playing agents in the latent space, reducing reliance on domain expertise and its associated biases. This latent space is then used to guide the exploration of diverse play styles within studied AI players.
\end{abstract}

\begin{IEEEkeywords}
play style, clustering, latent representation, autoencoder, unsupervised learning, deep learning, game analytics
\end{IEEEkeywords}

\section{Introduction}

Understanding play styles has numerous applications in game design, one of which is personalized game experiences. Dynamically adjusting game mechanics, difficulty, or content delivery based on a player's style can enhance engagement.

Style-coherent play data has been shown to improve game playing AI via imitation learning \cite{pearce_counter-strike_2021}. Additionally, it enables the creation of AI agents that exhibit diverse, human-like behaviors \cite{ingram_creating_2023}, \cite{mcilroy-young_learning_2022}.

Our main contribution is a high performance play style identification method that works with low-level
high-dimensional game trajectories, with results demonstrated on \textit{MicroRTS}.
This removes the need for handcrafted features and reduces reliance on domain expertise, allowing play styles to be identified directly from data. 
 This improves generalizability across different game environments.



\section{Related Work}
\subsection{Play Style}
Games span multiple disciplines, and as a result, the definition of play style varies across fields. For example, \cite{bartle_hearts_1996} proposes player archetypes based on their goals and interactions with other players and the environment. In contrast, \cite{yee_motivations_2006} applies factor analysis to player survey responses, incorporating in-game behavior, usage data, age, and gender.

Other studies align with the definition proposed by \cite{tychsen_defining_2008}, which categorizes play style within a three-tier hierarchy: play modes, play styles, and play personas. The work in \cite{bakkes_player_2012} identifies four approaches to player behavior modeling, distinguishing between models based purely on gameplay data and those grounded in psychology and sociology.

This research takes the quantitative approach from \cite{bakkes_player_2012}, unifying extracted patterns under the term \textit{play style}.

\subsection{Game Data Clustering}

One of the first large-scale player behavior studies, \cite{drachen_player_2009}, analyzed player profiles in \textit{Tomb Raider: Underworld} (TRU) using self-organizing maps to identify profiles based on player death statistics, time, and the use of the hint system. Additional clustering studies \cite{drachen_guns_2012}, \cite{drachen_guns_2016} have been conducted in \textit{Tera}, \textit{Battlefield 2}, and \textit{Destiny}. A common aspect of these studies is their reliance on aggregate statistics, which removes the sequential structure of play traces. In contrast, \cite{canossa_like_2018} identified common event subsequences in \textit{Tom Clancy’s The Division} and defined play styles based on their frequency.

\subsection{Deep Embeddings for Play Style Analysis}

Recently, the use of deep learning has been successful in learning latent representations in complex high-dimensional domains such as computer vision, natural language processing, and more, including video games.

A significant approach \cite{lin_unsupervised_2021} defines a metric to compare play traces based on shared deep encoded states and the actions made in response by players. In another relevant work \cite{ingram_play-style_2022}, the authors use embeddings from deep recurrent autoencoders to cluster play traces in \textit{Super Mario Bros}. Their sequence representations rely on handcrafted features such as the number of jumps in addition to action encodings.

\subsection{MicroRTS}
\textit{MicroRTS} is a lightweight RTS game environment used for AI research. It features typical challenges of RTS games such as combinatorial action space and real-time decision-making. Additionally, players need to manage units individually and simultaneously due to the lack of AI-assisted actions that are typical of commercial RTS games.

A yearly competition is held to develop AI agents to play \textit{MicroRTS}. As a result, strong AI agents have been developed, each exhibiting its own distinct play style.

\section{Methods}
\subsection{Data Description}
Our data is represented like \cite{huang_gym-rts_2021} with modifications: \textit{1)} We unify the direction parameters into a single direction parameter to reduce dimensionality without loss of information. \textit{2)} We use a two-dimensional representation of relative attack position to preserve distances. \textit{3)} We treat Hit Points, Resources, and Relative Attack Position as numerical variables.

Like \cite{huang_gym-rts_2021}, we represent observations in a game map of size $h\times w$ with a tensor of shape $(h,w,c_o)$ where each cell in the grid is represented by $c_o$ feature channels that represent the concatenation of the numerical features and the one hot encodings of the categorical variables. Similarly, the player actions will be represented with a tensor of shape $(h,w,c_a)$.

\subsection{Data Collection and Processing}

\begin{figure}
    \centering
    \includegraphics[width=0.66\linewidth]{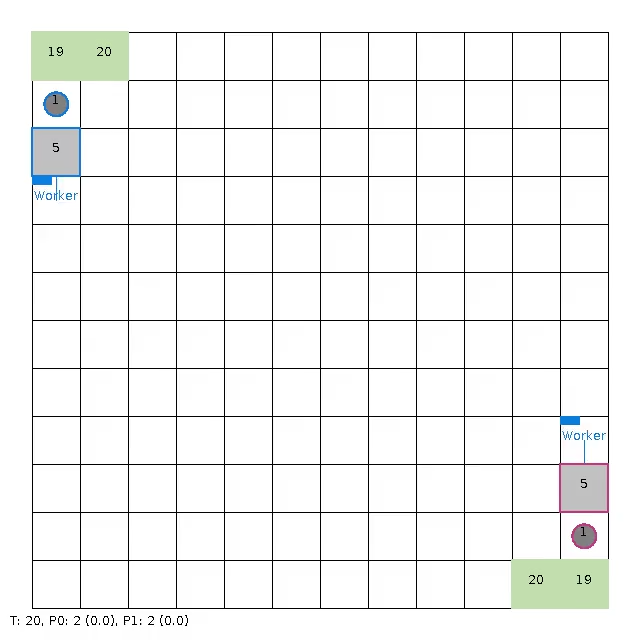}
    \caption{Example of \textit{MicroRTS} map. Players are represented by red and blue borders and start with a worker (grey circle), a base (grey square), and two resource mines (green squares).
}
    \label{fig:mapL}
\end{figure}

For our experiments, we record games played using 13 AI agents (PassiveAI,	RandomBiasedAI,	RandomAI, WorkerRush,	LightRush, NaiveMCTS, MixedBot, Rojo, Izanagi, mayari, Tiamat,	Droplet, and	CoacAI). Games are played in maps of the family \verb|12x12basesWorkers{i}| where \verb|i| $\in\{$\verb|A,B,C,D,E,F,G,H,I,J,K,L|$\}$. See Fig. \ref{fig:mapL} for map \verb|L|, other maps will be variations of the starting placement of units. Each match-up is played 10 times, including self-play match-ups and we record the two points of view (POV) of each game. The two extracted POVs will be centered around player 1, that is, we swap the owner labels for the POV of player 2. We collect states and actions for time steps where the player issued an action to compress the data.

We use the play traces from maps \verb|A| to \verb|K| for training and validation, 2 play traces from each match-up for validation, 8 for training. The test set is comprised of play traces played in map \verb|L|. Numerical data is scaled to the $[0,1]$ range and data is augmented by randomly mirroring vertically and/or horizontally. We do not augment validation or test data. 

From each sequence we extract subsequences\footnote{Going forward we will use sequence to refer to these subsequences.} of length $32$ with a stride of $8$ for training, meaning there will be overlapping subsequences, and $32$ for testing and validation.

\subsection{Model Architecture}
\begin{figure*}
    \centering
    \includegraphics[width=0.66\linewidth]{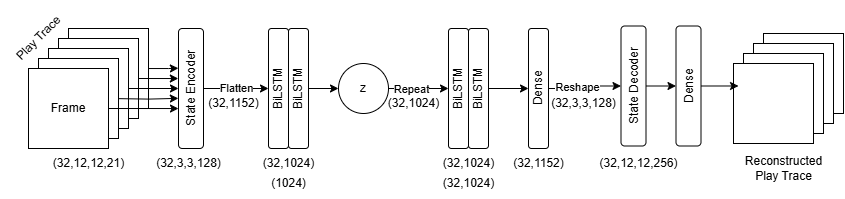}
    \caption{Model architecture for the actions-only autoencoder with output sizes. Batch size is omitted for clarity. }
    \label{fig:arch}
\end{figure*}

We train models to encode sequences of different types of information: actions, states, and a joint model for both. We will use frame to refer to a time slice of actions, states, or both. To encode our player trajectories, we use an autoencoder architecture based on the LSTM-autoencoder from \cite{ingram_play-style_2022} combined with the encoder-decoder from \cite{huang_gym-rts_2021}. See Fig. \ref{fig:arch}.

\subsubsection{Encoder}
 We first encode each frame using a frame encoder network shared for all frames, meaning it will be time-agnostic. This frame encoder is a standard convolutional neural network (CNN) with two blocks of Convolution, Max Pooling, and GELU. 
The result of this is a sequence of encoded frames that will be further encoded by two stacked Bidirectional Long Short-Term Memory (BiLSTM) layers into the latent representation $z$ of size 1024 (512 from each LSTM direction).

\subsubsection{Decoder}
The latent vector will be passed as input to 2 BiLSTM layers as many times as the original sequence length. The result of this will be passed to a dense layer to the appropriate dimension to be reshaped and upsampled with a transpose CNN for each output in the sequence. This transpose CNN or frame decoder will also be time agnostic, like the frame encoder. Finally, the output is passed to the final classifier layer to obtain distribution probabilities for categorical features and predicted values for numerical features.

As a baseline to compare results, we use aggregate action measures to define handcrafted features for each play trace: for each categorical feature, we use the frequency of each non-null class among non-null classes for the same categorical variable and the sum of relative attack positions $x$ and $y$. This results in 18 handcrafted features for each sequence.

\subsection{Clustering}

Once the autoencoder has been trained, the data is encoded into the latent space using the encoder. We use Principal Component Analysis (PCA) to reduce the latent representations to a lower-dimensional 64D vector before clustering with $k$-means. For the handcrafted features baseline, we apply clustering on the defined 18 features directly. We limit the clustering to sequences that are coherent both temporally (e.g., the starting subsequence) and spatially (same map and same side), following previous results showing that styles are dependent on game levels, modes, and maps \cite{sifa_behavior_2013-1},\cite{ravari_playing_2018}.

Visualizations use a t-distributed stochastic neighbor embedding (t-SNE) of the PCA vectors. An interactive tool is created to explore the t-SNE space, where points in the space can be selected to render a video of the play trace assigned to that point in the t-SNE.

Regarding the choice of the number of clusters, the decision is not trivial. This is especially true in the true unsupervised regime, where there are no ground truth labels. While there are different approaches to do this for each clustering, to allow for comparable results across maps and input schemes we evaluate three choices of number of clusters $k$: 10, 13, and 16.

Because different AI agents are being used to generate the play traces, we can assign the agent name as the ground truth label. Note how this is not a play style label, some agents may change their behaviour depending on the opponent, while some may be similar in play style. We track completeness and homogeneity to allow for these situations. Additionally, we track Adjusted Rand Index (ARI) and Adjusted Mutual Information (AMI), which are measures of how well the original ground truth labels are recovered.

\section{Results and Discussion}

\subsection{Clustering}
We present results for the starting subsequence of every match in Table \ref{tab:combined_clustering_metrics}, we average the results for all maps (\verb|A|-\verb|K|) and also present individual results for \verb|L|.

\begin{table}
    \centering
    \setlength{\tabcolsep}{4pt} 
    \cprotect\caption{Clustering metrics: average for maps \verb|A|-\verb|K|, single map \verb|L|. (E)mbedding: (S)tates, (A)ctions, (J)oint, (H)andcrafted}
    \renewcommand{\arraystretch}{1.2}
    \begin{tabular}{|l|c|cc|cc|cc|cc|}
        \hline
        \multirow{2}{*}{E} & \multirow{2}{*}{$k$} 
        & \multicolumn{2}{c|}{Completeness} & \multicolumn{2}{c|}{Homogeneity} 
        & \multicolumn{2}{c|}{ARI} & \multicolumn{2}{c|}{AMI} \\
        \cline{3-10}
        & & \verb|A|-\verb|K| & \verb|L| & \verb|A|-\verb|K| & \verb|L| &\verb|A|-\verb|K| & \verb|L| &\verb|A|-\verb|K| & \verb|L|    \\
        \hline
        S         & 10 & 0.666 & 0.651 & 0.540 & 0.478 & 0.354 & 0.272 & 0.590 & 0.544 \\
        S         & 13 & 0.652 & 0.622 & 0.601 & 0.566 & 0.422 & 0.407 & 0.618 & 0.585 \\
        S         & 16 & 0.641 & 0.615 & 0.631 & 0.612 & 0.423 & 0.411 & 0.627 & 0.605 \\\hline
        J  & 10 & 0.696 & \textbf{0.788} & 0.543 & 0.630 & 0.330 & 0.429 & 0.603 & 0.696 \\
        J  & 13 & 0.691 & 0.751 & 0.630 & 0.692 & 0.440 & 0.516 & 0.652 & \textbf{0.715} \\
        J  & 16 & 0.665 & 0.705 & 0.651 & \textbf{0.713} & 0.447 & \textbf{0.519} & 0.649 & 0.703 \\\hline
        A          & 10 & \textbf{0.736} & 0.776 & 0.597 & 0.628 & 0.412 & 0.447 & 0.653 & 0.690 \\
        A          & 13 & 0.704 & 0.709 & 0.634 & 0.643 & 0.422 & 0.442 & \textbf{0.660} & 0.668 \\
        A          & 16 & 0.690 & 0.695 & \textbf{0.681} & 0.713 & \textbf{0.468} & 0.498 & 0.677 & 0.698 \\\hline
        H          & 10 & 0.649 & 0.715 & 0.538 & 0.610 & 0.401 & 0.457 & 0.582 & 0.653 \\
        H          & 13 & 0.636 & 0.647 & 0.577 & 0.609 & 0.421 & 0.417 & 0.597 & 0.620 \\
        H          & 16 & 0.638 & 0.674 & 0.625 & 0.691 & 0.452 & 0.507 & 0.622 & 0.675 \\\hline
    \end{tabular}
    \label{tab:combined_clustering_metrics}
\end{table}
Regarding the general performance of the different embeddings, the actions-only and the joint encoders edge ahead of the states-only encoder and the handcrafted features. The joint model produces clusterings that are competitive with those from the actions-only model across maps, taking the lead in some of the clustering metrics. When comparing maps, better clustering metrics are observed across all embeddings except states-only, suggesting that map \verb|L| might be intrinsically easier to cluster, causing agents to play with more distinct styles.

We hypothesize that the use of states-only is not well suited for our specific problem as it uses a single feature (owner) to break the symmetry. Meanwhile, the other three encodings contain information regarding the actions of only one of the players, resulting in encodings more closely related to the player that produced the play trace.

\subsection{Qualitative Exploration}

\begin{figure*}
    \centering
    \includegraphics[width=\linewidth]{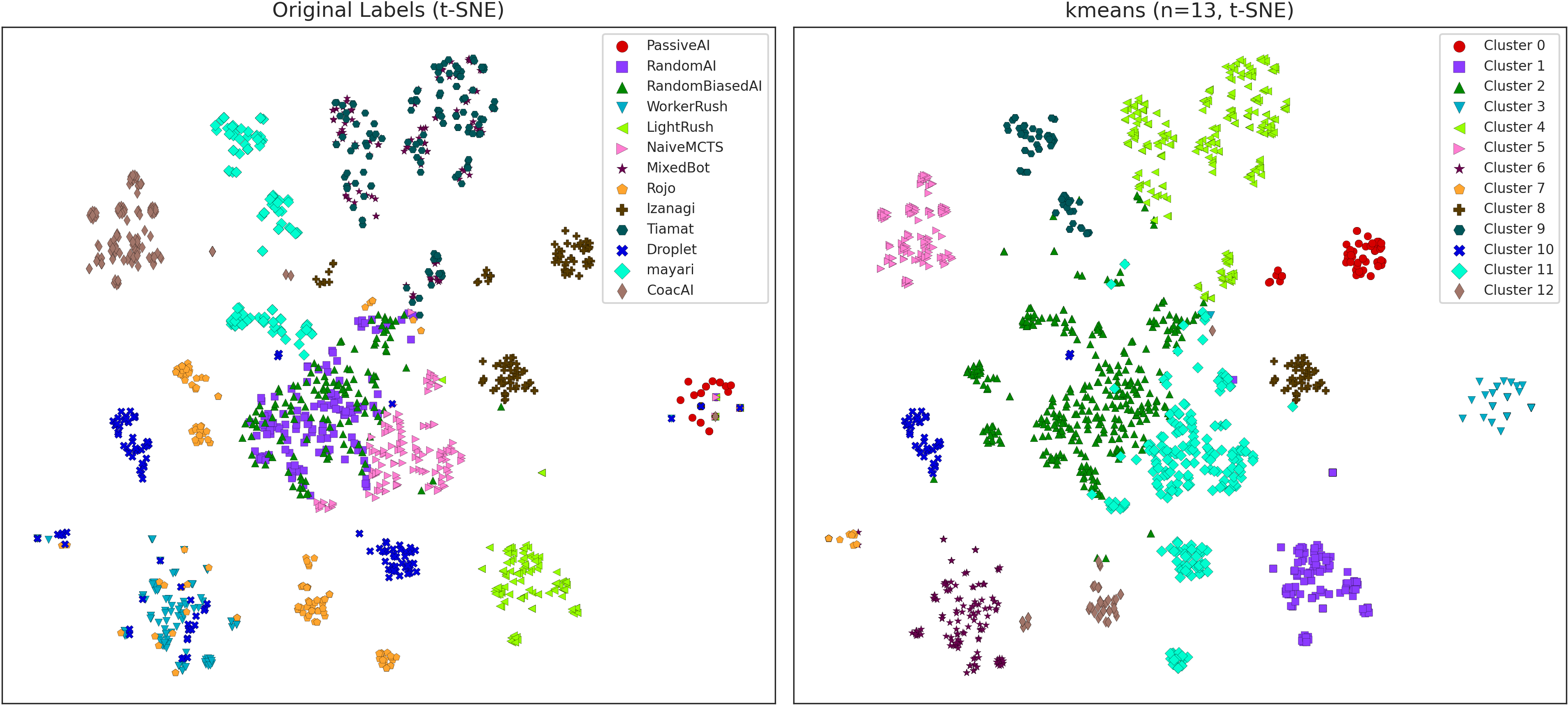}
    \cprotect\caption{t-SNE for joint model embeddings of map \verb|L| games with original labels (left) and 13 $k$-means clusters (right). Best visualized in the electronic version.}
    \label{fig:tsne}
\end{figure*}

To validate and visualize our results, we present the t-SNE from the best latent representation for play traces of map \verb|L| in terms of AMI, see Fig. \ref{fig:tsne}.

Notably, the clustering algorithm merges the play traces from MixedBot and Tiamat. This is to be expected as MixedBot is an ensemble of agents, one of which is Tiamat. A small group of these two agents is clustered separately with other types of agents. Upon inspection, this cluster consists of agents that were defeated by a WorkerRush-like strategy, which uses workers to attack and overwhelm the agent before it can build units, meaning they did not have the time to show their playstyle before being defeated. 

We now shift our attention to play traces from mayari, one of the strongest agents. Most of its traces are clustered together showcasing optimized resource gathering with a quick production of heavy units and a balance of production and attack with workers. However, some of its traces are clustered together with the random variants which should not happen with such a strong agent. Inspecting these play traces, we see that depending on which worker is chosen to build the barracks, the other worker's path to the resources can end up being disrupted, resulting in erratic behavior where it looks like it is moving back and forth. This behavior is later corrected outside of our chosen subsequence length.

Droplet has been described as a WorkerRush strategy enhanced with Monte Carlo Tree Search (MCTS) \cite{huang_gym-rts_2021}, which lines up with it having a portion of its traces clustered with the WorkerRush agent and another portion grouped with the NaiveMCTS agent. We also see a few agents that are clustered together with the PassiveAI, these are play traces of agents that failed to run or crashed during the data collection process.


The 2D t-SNE representation can be deceiving: for example, one may think that there is one big cluster of the random variants and the NaiveMCTS traces, however, the $k$-means clustering separates the NaiveMCTS from the random variants quite successfully.


\section{Conclusion}

This work explores the use of CNN-LSTM autoencoders to learn embeddings of low-level representations of play traces to identify play styles, reducing reliance on domain expertise during feature creation. By using raw action and state sequences instead of handcrafted features, our approach minimizes domain-specific biases, allowing patterns to emerge naturally from the data. This enables adaptability across different games and ensures that discovered play styles are determined by player behavior rather than predefined assumptions on style-defining features.

Additionally, the obtained latent space can be used to guide the analysis of the play traces. We perform a qualitative analysis of the obtained latent space and identify different play patterns within players, including erratic behaviors showing specific failings of the studied AI agents.

A natural extension for this study would be to apply it to a natural domain, such as modern commercial games. On the embedding side, future work should focus on autoencoder structures that induce specific latent structures like VAEs or VQ-VAEs, and the use of transformers to scale to bigger datasets and higher dimensions required to analyze the player bases of modern games.

\bibliographystyle{IEEEtrannourl}
\bibliography{references}

\end{document}